\documentclass[PHME, 2022]{PHMSociety}

\usepackage{graphicx}
\graphicspath{{Figures/}} 

\usepackage{amsmath,amssymb,amsfonts}
\usepackage{algorithmic}
\usepackage{enumerate}
\usepackage[ruled,vlined]{algorithm2e}
\usepackage{siunitx}
\SetArgSty{}

\usepackage{graphicx}
\usepackage{textcomp}
\usepackage{xcolor}
\usepackage{booktabs}

\usepackage{microtype}

\begin{document}
\sloppy
\title{Autoencoder based Anomaly Detection and Explained Fault Localization in Industrial Cooling Systems}

\author{%
	Stephanie Holly \authorNumber{1,2}, Robin Heel \authorNumber{1}, Denis Katic \authorNumber{3}, Leopold Schoeffl \authorNumber{4}, Andreas Stiftinger 
	\authorNumber{4}, Peter Holzner
	\authorNumber{1,2}, Thomas Kaufmann
	\authorNumber{1,2}, Bernhard Haslhofer
	\authorNumber{3}, Daniel Schall \authorNumber{1}, Clemens Heitzinger \authorNumber{2}, and Jana Kemnitz \authorNumber{1}
}

\address{
	\affiliation{{1}}{Siemens Technology, Vienna, 1210, Austria}
	\tabularnewline 
	\affiliation{2}{Vienna University of Technology, Vienna, 1040, Austria}{ 
		}
	\tabularnewline
	\affiliation{3}{Austrian Institute of Technology, Vienna, 1210, Austria}
	\tabularnewline
	\affiliation{4}{Hauser, Linz, 4040, Austria}
}

\maketitle
\pagestyle{fancy}
\thispagestyle{plain}

\phmLicenseFootnote{Stephanie Holly}

\begin{abstract}%

Anomaly detection in large industrial cooling systems is very challenging due to the high data dimensionality, inconsistent sensor recordings, and lack of labels. The state of the art for automated anomaly detection in these systems typically relies on expert knowledge and thresholds. However, data is viewed isolated and complex, multivariate relationships are neglected. In this work, we present an autoencoder based end-to-end workflow for anomaly detection suitable for multivariate time series data in large industrial cooling systems, including explained fault localization and root cause analysis based on expert knowledge. We identify system failures using a threshold on the total reconstruction error (autoencoder reconstruction error including all sensor signals). For fault localization, we compute the individual reconstruction error (autoencoder reconstruction error for each sensor signal) allowing us to identify the signals that contribute most to the total reconstruction error. Expert knowledge is provided via look-up table enabling root-cause analysis and assignment to the affected subsystem. We demonstrated our findings in a cooling system unit including 34 sensors over a 8-months’ time period using 4-fold cross validation approaches and automatically created labels based on thresholds provided by domain experts. Using 4-fold cross validation, we reached a F1-score of 0.56, whereas the autoencoder results showed a higher consistency score (CS of 0.92) compared to the automatically created labels (CS of 0.62) -- indicating that the anomaly is recognized in a very stable manner. The automatically created labels, however, detected anomaly earlier. The main anomaly was found by the autoencoder and automatically created labels, and was also recorded in the log files. Further, the explained fault localization highlighted the most affected component for the main anomaly in a very consistent manner.
\end{abstract}

\section{Introduction}
\label{intro} 
Malfunctions or even a failure of refrigeration systems are a risk with very high damage potential for food wholesalers. In the course of Industry 4.0 and digitization, sensors and instrumentation drive the central forces of innovation. New potentials for monitoring and machine learning based predictive maintenance of the cold stores are opening up. However, the training and updating of such machine learning based models poses several challenges. First, damage and outage reports, which represent the required ground-truth for a supervised learning task, are not yet collected in a systematic and consistent manner. Second, a refrigerating system is a large, complex system including several hundreds of sensors presenting a widely varying data domain such as temperature, vibration or engine speed \cite{Weerakody}. Third, the sensors are often retrofitted or upgraded successively in the course of digitization. Therefore the question remains open how a machine learning model can be scaled from one component to an entire system or several systems.

The challenge of missing ground truth data and therefore learning useful representations with little or no supervision is a key challenge in machine learning. In the context of predictive maintenance, unsupervised learning has shown to be successful in identifying system failures without supervision. However, identifying the root cause of the failure employing unsupervised learning procedures remains an open task. For this reason we transform our unsolvable task into a closely related solvable task, which aims to achieve a similar benefit and business impact. Therefore, we have started with a requirements analysis. The outcomes of the requirements analysis showed us that the localization of the affected sensors, associated with the root cause of the detected failures are an important step. As the affected system is very complex including several hundreds of sensors, the maintenance employee can save valuable time and effort if the affected sensors and thereby the affected subsystem can be localized. Therefore we propose an algorithm for fault localization in large cooling systems with no supervision or little supervision.

\begin{itemize}
    \item[\textbf{C1:}] We define a real-world learning task based on industrial requirements and provide a $18$ months ground truth data set for an entire cooling system unit including $34$ sensor signals.
    \item[\textbf{C2:}] We provide an autoencoder based anomaly detection workflow suitable for multivariate and increasingly upgraded time series data.
    \item[\textbf{C3:}] Our workflow includes an algorithm for explained fault localization based on the individual reconstruction error for each sensor signal.
    \item[\textbf{C4:}] Our workflow includes a root cause analysis enabled by integrated expert knowledge.
    \item[\textbf{C5:}] Our workflow was compared against automatically created labels showing an F1-score of $0.56$ and a consistency score of $0.92$. The explained fault localization highlighted the most affected component in a very consistent manner.
     

     
\end{itemize}

\section{Related Work}
\label{related-work}

Industrial cooling systems (ICS) are widely deployed in large supermarkets and storage warehouses to preserve perishables with a global market valued to over USD 5 billion \cite{Report2020}. ICS are subject to faults, such as compressor failures and bearing damage, that can degrade the operational efficiency and even result in their breakdown. Accurate and timely detection of faults and degradation is critical to prevent food spoilage, customer inconvenience, maintenance costs, and other related losses. Automated fault diagnosis in ICS has been explored for many years \cite{GRIMMELIUS1995}, ranging from Kalman-filter based methods \cite{YANG2011}, random forest \cite{Kulkarni2018} and neural network based approach. AI based predictive maintenance is estimated to decrease breakdowns by up to 70 \% and lowers maintenance costs by 25 \% \cite{Deloitte2017} in the next years. 

Prognostics and health management approaches have been studied extensively across industrial applications, such as aircraft systems \cite{Bieber2021}, hard drives \cite{Barelli12021}. While some traditional methods for fault detection require feature engineering \cite{kato2001integrating, Su2019, Wang2020}, recent work has shown that end-to-end autoencoders can outperform traditional approaches \cite{zong2018deep, MALEKI2021, heistracher2021}. Autoencoders are promising and have been for minimal-configuration fault detection \cite{heistracher2021, Hood2021}, however this unsupervised methods lack the ability for fault classification or fault location. 

Fault classification is typically based on supervised learning \cite{IsmailFawaz2019, Dempster2020, Wang2017, Karim2018} and required a large number of ground truth data, often lacking in industrial applications. Furthermore, these supervised approaches are often difficult to scale and difficult to transfer to other, similar systems \cite{kemnitz2021, Heistracher2021a}. As the affected system is very complex including several hundreds of sensors, the maintenance employee can save valuable time and effort if the affected sensors and thereby the affected subsystem can be localized. Further, explaining and quantifying the individual contribution helps increase trust and interpretability \cite{GREZMAK2019}. The knowledge about individual localization and contribution can be combined with expert knowledge and thereby enables root cause analysis.

Therefore we propose an end-to-end autoencoder based workflow for fault localization in large cooling systems with no supervision or little supervision. This proposed workflow is inspired by heat and saliency maps \cite{simonyan2014} applied in computer vision \cite{Goebel2018}. 

\section{Industrial Requirements}
Hauser aims to build a machine learning based monitoring, alarm, and remote maintenance systems for industrial refrigeration systems, which are deployed in hundreds of locations around the world. Since many of these systems are of the same type, a model-driven  approach that could predict damages or outages to cold storage's, would scale from a business perspective and could also be offered as a service to customers. The following requirements result from this vision:

\begin{itemize}
    \item[R1:] Employable without ground truth data
    \item[R2:] Scalable to similar refrigeration systems
    \item[R3:] A holistic approach employing a cockpit view
    \item[R4:] Employable for fault detection
    \item[R5:] Employable for fault localization
    \item[R6:] Data agnostic
    \item[R7:] Dealing with retrofitted or upgraded sensors
\end{itemize}

%
%
%

It is important to start with an approach that is expandable. Each step should add substantial business value. 


\section{Machine Learning Workflow} \label{Machine Learning Method}

\subsection{Overall Workflow} \label{Overall Workflow}

\begin{figure}[ht]
	\centerline{\includegraphics[width=0.4\textwidth]{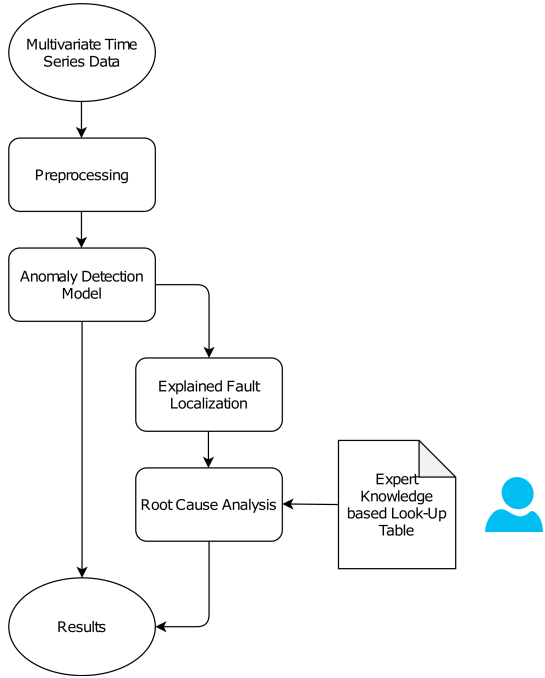}}
	\caption{Overall workflow: preprocessing, anomaly detection, explained fault localization, and root cause analysis.}
	\label{overall workflow}
\end{figure}

We propose a workflow for preprocessing, anomaly detection, explained fault localization and root cause identification of multivariate time series data, see Fig. \ref{overall workflow}. For anomaly detection, we use a LSTM autoencoder. We identify system failures using a threshold on the total reconstruction error $RE_{\mathrm{total}}(S)$ Eq. (\ref{totalrecon}) of sensor signals $S:=[S_{1}, \dots, S_{n}]$. For fault localization, we compute the individual reconstruction error $RE_{\mathrm{ind}}(S_{i})$ Eq. (\ref{indrecon}) for each sensor signal $S_{i}$ allowing us to identify the signals that contribute the most to the total reconstruction error $RE_{\mathrm{total}}(S)$. Provided an expert knowledge "look-up table", we can thus locate the affected subsystem in the cooling system.

\subsection{Preprocessing} \label{Preprocessing}

Data preprocessing is a crucial task in machine learning pipelines \cite{JoergLeukel}. Here, data preprocessing denotes the process of preparing raw data for the machine learning model including data cleaning, signal selection, resampling, missing values treatment and data normalization. In industrial systems, data preprocessing is a major challenge due to its high dimensionality and complex structure \cite{EbruBekar}. In industrial cooling systems, we find numerous components with various data sources including several hundreds of sensors, a widely varying data domain, and retrofitted and upgraded sensors requiring distinct approaches in the mentioned preprocessing steps.

In the following, we want to present our approach to data preprocessing in an industrial cooling system with time series data. Let $S_{i}$ denote the $i^{\text{th}}$ sensor signal in the cooling system, $1 \le i \le n$. The data is acquired without a defined frequency, that is, the data is not available at the same frequency and each sensor signal $S_{i}$ is recorded with an individual frequency or irregularily. 

\subsubsection{Data Cleaning} \label{Data Cleaning}
In the course of digitization, sensors are often retrofitted or upgraded successively. We therefore removed early time periods lacking sensor signals considered important by Hauser experts.

\subsubsection{Signal Selection} \label{Signal Selection}
The industrial cooling system consists of numerous components with a wealth of data sources providing a huge amount of sensor signals. The massive amount of data is a key challenge in data preprocessing \cite{EbruBekar}. In order to decrease the number of signals, we calculated the correlation coefficient and removed highly correlated signals \cite{Nahian}. Let $X$ and $Y$ denote random variables with covariance $\operatorname{cov}(X, Y)$ and standard deviation $\sigma_{X}$, $\sigma_{Y}$ respectively. Then, the correlation coefficient $\rho_{X, Y}$ is given by
\begin{align}
    \rho_{X, Y} := \frac{\operatorname{cov}(X, Y)}{\sigma_{X}\sigma_{Y}}.
\end{align}

Extracting information from the timestamp can increase the quality of prediction models \cite{EvgeniyLatyshev}. Thus, we added the signals month, time (hour) and weekday to our observations.

\subsubsection{Resampling} \label{Resampling}

The acquired data consists of irregularly sampled time series data. With the growth of multi-sensor systems, the preprocessing of irregular time series data is becoming increasingly important \cite{Weerakody}. Due to numerous components with various data sources including several hundreds of sensors and a widely varying data domain, we cannot expect to find all sensor signals sampled at a constant sampling rate with common timestamps. For modeling, however, we need to resample the data to a regular frequency. Let $x_{t}^{i}$ denote a observed value of sensor signal $S_{i}$ at time $t$. Let $r$ denote a regular sampling rate. We then obtain new timestamps $\mathcal{T}$ by equidistant time intervals of length $r$. Depending on the signal type, we compute a new value $\hat{x}_{t}^{i}$ of signal $S_{i}$ at timestamp $t \in \mathcal{T}$. 
In general, the acquired sensor signals represent numerical values, e.g.\ measuring temperature, vibration or engine speed. However, the cooling system also includes signals of boolean values giving information about the system's health and up-counting signals giving information about the pause time of components in the system. When the value is counting, the minimum value is the most representative one, as it was the initial position of the counter. And using the maximum boolean value, was the same result as using the logical OR, which means that it is zero if and only if all samples of the corresponding interval are zero.
In the general case, we simply average over all observed values $x_{u}^{i}$ in the interval $\big[t, t+r \big)$, that is,
\begin{align}
    \hat{x}_{t}^{i}:=\frac{1}{\# \big \lbrace x_{u}^{i} |  u \in \big[t, t+r \big) \big \rbrace} \underset{u \in \big[t, t+r \big)}{\sum} \hspace{1mm} x_{u}^{i}.
\end{align}
In the case of boolean values $x^{i} \in \lbrace 0, 1 \rbrace$, we define
\begin{align}
    \hat{x}_{t}^{i}:=\underset{u \in [t, t+r)}{\max} \hspace{1mm} x_{u}^{i},
\end{align}
and in the case of constantly (by $c \in \mathbb{N}$) increasing values $x^{i} \in \lbrace k + c | k \in \mathbb{N} \rbrace$, we define
\begin{align}
    \hat{x}_{t}^{i}:=\underset{u \in [t, t+r)}{\min} \hspace{1mm} x_{u}^{i}.
\end{align}

\subsubsection{Missing Values Treatment} \label{Missing Values Treatment}
Resampling irregularly sampled time series data will result in missing values for one or more sensor signals $S_{i}$ at a given timestamp $t \in \mathcal{T}$. Simple statistical techniques include forward-filling and zero imputation \cite{Weerakody}. We used the fill-forward method for missing values. Before the first appearance of a value, we initialized a default value by computing the median of all available values.

\subsubsection{Feature Representation} \label{Feature Engineering}
In contrast to the acquired signals in the cooling system, let features define the representation of data feed into the machine learning model. Let $w$ denote the window size. Applying a window of size $w$ to the sensor signals $S_{i}$ at timestamp $t$, the features $F^{t}$ at timestamp $t$ are given by
\begin{align}
    F^{t} := \begin{pmatrix}
    \hat{x}^{1}_{t} &\hat{x}^{2}_{t} & \hdots & \hat{x}^{n}_{t} \\
    \hat{x}^{1}_{t+1} & \hat{x}^{2}_{t+1} & \hdots & \hat{x}^{n}_{t+1}\\
    \vdots & \vdots &   & \vdots\\
    \hat{x}^{1}_{t+w-1} & \hat{x}^{2}_{t+w-1} & \hdots & \hat{x}^{n}_{t+w-1} \end{pmatrix}  \in \mathbb{R}^{w \times n}.
\end{align}

For modeling, we reshaped the features and obtained
\begin{align}
    F^{t} = \begin{pmatrix}
    F_{1} \\
    F_{2}\\
    \vdots \\
    F_{w n}
    \end{pmatrix}  \in \mathbb{R}^{w n},
\end{align}
where $F_{j}$ with $j = i  w + k$ corresponds to $\hat{x}^{i}_{t+k}$.

\subsubsection{Data Normalization} \label{Data Normalization}
Using min-max normalization, we standardized the features by scaling each feature individually to the range $(0, 1)$. The training data is scaled, and then the scaling parameters are applied to the test data. In the domain of machine learning, normalization plays a key role in the preprocessing of data including variables of different scale. In normalization, each variable is scaled individually to the range $(0, 1)$ avoiding a variable dominating the machine learning model.

\subsection{Anomaly Detection Model} \label{Anomaly Detection Model}

\begin{figure}[ht]
	\centerline{\includegraphics[width=0.13\textwidth]{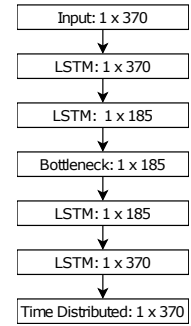}}
	\caption{Model Architecture}
	\label{model-architecture}
\end{figure}

We applied a LSTM autoencoder with the following settings:
an input layer of size $1 \times 370$ ($37$ signals and a window of size $10$), the first encoding layer with output size $1 \times 370$, the second encoding layer with output size $1 \times 185$, a repeat vector with output size $1 \times 185$, the first decoding layer with output size $185$, the second decoding layer with output size $370$, and a time distributed layer of size $1 \times 370$, see Fig. \ref{model-architecture}.

For hyper-parameter tuning, we used Bayesian optimization. Turner et al. \cite{Turner} demonstrated decisively the benefits of Bayesian optimization over random search and grid search for tuning hyperparameters of machine learning models. Bayesian optimization benefits from previous evaluations of hyper-parameter configurations by including past hyper-parameter configurations in the decision of choosing the next hyper-parameter configuration. Therefore, it avoids unnecessary evaluations of the expensive objective function and requires fewer iterations to find the best hyper-parameter configuration.


We searched for the sampling rate, number of layers, dropout rate, activation function, optimizer, learning rate and batch size. We obtained a sampling rate of $60$ seconds, $2$ autoencoder layers, a dropout rate of $0.2$, the $\tanh$ activation function, the rmsprop optimizer, a learning rate of $0.001$, and a batch size of $16$.

\begin{figure}[ht]
	\centerline{\includegraphics[width=0.45\textwidth]{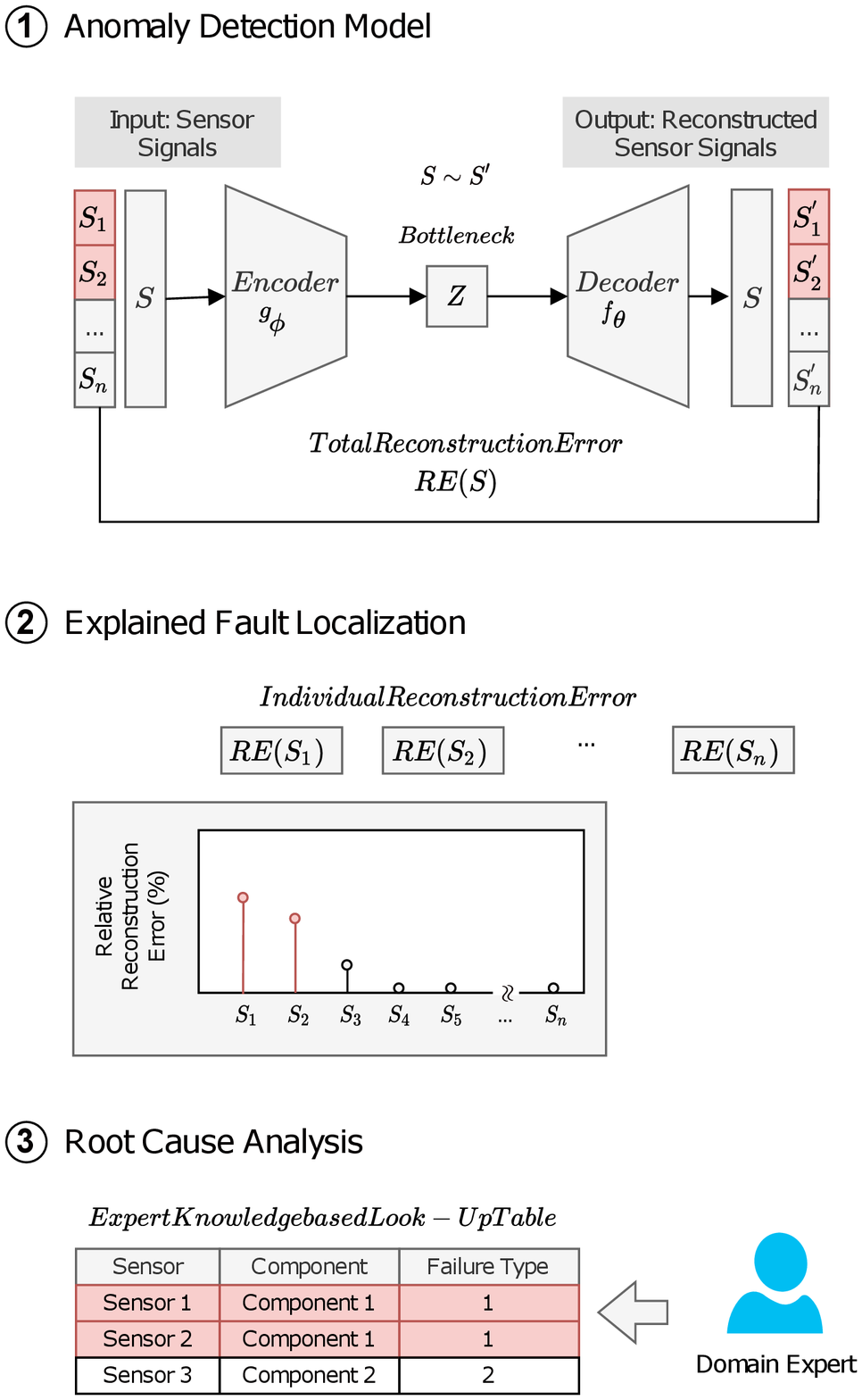}}
	\caption{Graphical Abstract of Paper, including 1) anomaly detection (identify system failures using a threshold on the total reconstruction error of all sensor signals), 2) explained fault localization (compute the the individual reconstruction error for each sensor signal and identify affected signals), 3) and root cause analysis (locate the affected subsystem based on a look-up table and the affected signals).}
	\label{graphical-abstract}
\end{figure}

\subsection{Explained Fault Localization} \label{Fault Localization}

We propose an end-to-end autoencoder
based workflow for fault localization in a large cooling system not providing ground truth data. The total reconstruction error is given by Eq. (\ref{totalrecon}). Motivated by heat and saliency maps \cite{simonyan2014} applied in computer vision \cite{Goebel2018}, we derive from the individual reconstruction error Eq. (\ref{indrecon}) how much each sensor signal $S_{i}$ contributes to the total reconstruction error Eq. (\ref{totalrecon}). Thus the individual reconstruction error Eq. (\ref{indrecon}) can be used to identify affected signals and thereby locate the affected subsystem. 

The total reconstruction error of feature $F^{t}$ at timestamp $t$ is given by 
\begin{align}\label{totalrecon}
    RE_{\mathrm{total}}(F^{t}) := \frac{1}{w n} \sum_{j=1}^{w n} (F_{j} - F^{\prime}_{j})^{2},
\end{align}
where 
\begin{align}
F^{\prime t} := \begin{pmatrix}
    F^{\prime}_{1} \\
    F^{\prime}_{2}\\
    \vdots \\
    F^{\prime}_{w n}
    \end{pmatrix} \in \mathbb{R}^{w n}
\end{align} is the prediction of feature $F^{t}$ at timestamp $t$.

We define the individual reconstruction error of signal $S_{i}$, $1 \le i \le n$,  at timestamp $t$ by

\begin{align}\label{indrecon}
    RE_{\mathrm{ind}}(S_{i}) := \frac{1}{w} \sum_{k=0}^{w-1} (\hat{x}_{t+k}^{i} - \hat{x}^{\prime i}_{t+k})^{2},
\end{align}
where $\hat{x}^{i}_{t+k}$ and $\hat{x}^{\prime i}_{t+k}$ correspond to $F_{j}$ and $F^{\prime}_{j}$ with $j = i  w + k$ respectively. 

In a $k$-fold cross validation approach, we compute for each test dataset $l$, $1 \le l \le k$, a threshold 
\begin{align}\label{threshold}
    T_{l} := \mu_{l} + c \sigma_{l},
\end{align}
where $\mu_{l}$ and $\sigma_{l}$ are the mean and the standard deviation of the total reconstruction error $RE_{\mathrm{total}}$ on the remaining training data and $c$ is a constant found by plotting ROC curves, see Fig.~\ref{ROC}. 

In algorithm \ref{alg: Fault Localization}, we give the pseudocode for the Fault Localization algorithm. The algorithm takes the threshold $T$, the feature $F^{t}$ at timestamp $t$, the predicted feature $F^{\prime t}$ at timestamp $t$ and the number of significant signals $m$, $1 \le m \le n$, as an argument, and returns either the $m$ most significant signals $S^{*}$ or the empty set. We compute the total reconstruction error $RE_{\mathrm{total}}$ of feature $F^{t}$ at timestamp $t$ based on Eq. (\ref{totalrecon}). If the total reconstruction error $RE_{\mathrm{total}}$ exceeds the threshold $T$, we determine the $m$ most significant signals $S^{*} \subset \lbrace S_{1}, S_{2}, \dots , S_{n} \rbrace$. Therefore, we compute the individual reconstruction error $RE_{\mathrm{ind}}[i]$ based on Eq. (\ref{indrecon}), $i=1 \dots n$. Then, we select iteratively the signal $idx$ that yields the highest individual reconstruction error $RE_{\mathrm{ind}}[idx]$, append signal $idx$ to $S^{*}$ and remove the selected signal from further calculations by setting its value to zero. The method APPEND takes a set and an element as an argument and appends the element to the set. When the iteration terminates, set $S^{*}$ contains exactly the $m$ most significant signals, that is, the signals with the highest individual reconstruction error. If the total reconstruction error $RE_{\mathrm{total}}$ does not exceed the threshold $T$, $S^{*}$ is the empty set. Finally, the set $S^{*}$ is returned.

\begin{algorithm}\small
	\DontPrintSemicolon
	\caption{Fault Localization}
	\label{alg: Fault Localization}
	\SetAlgoLined
	\textbf{Input:} threshold $T$, feature $F^{t}$ at timestamp $t$, predicted feature $F^{\prime t}$ at timestamp $t$, number of significant signals $m$
	
    \textbf{Output:} significant signals $S^{*} \subset \lbrace S^{1}_{t}, S^{2}_{t}, \dots , S^{n}_{t} \rbrace$ with $\# S^{*} = m$
    
    $RE_{\mathrm{total}} := \frac{1}{w n} \sum_{j=1}^{w n} (F_{j} - F^{\prime}_{j})^{2}$
    
    \eIf { $RE_{\mathrm{total}} > T$ }{

        $RE_{\mathrm{ind}}[i] := \frac{1}{w} \sum_{k=0}^{w-1} (\hat{x}_{t+k}^{i} - \hat{x}^{\prime i}_{t+k})^{2}$ \hspace{2mm} $i=1,\dots,n$

	    $S^{*} := \emptyset$
	    
	    \For{$i=1,\dots,m$}{
            $idx := \underset{i=1, \dots, n}{\arg \max} \hspace{1mm} RE_{\mathrm{ind}}[i]$\;
            APPEND$(S^{*}, idx)$\;
            $RE_{\mathrm{ind}}[idx] := 0$
	    }
        
        }{
            $S^{*} := \emptyset$
        }
        
    return $S^{*}$
\end{algorithm}

\subsection{Root Cause Analysis and Integrated Expert Knowledge} \label{Root Cause Analysis}

The proposed workflow for fault detection and root cause identification -- the identification of the affected sensor signals $S_{i}$ -- allows us to locate the affected subsystem. Each sensor signal $S_{i}$ is assigned to a component in the cooling system. The component can then be used to determine the root-cause and failure type. Domain knowledge and physical connections are stored in the system using a look-up table. Over the years, the experts have collected which sensors are typically associated with a root-cause. In our case,  a physics-aware look up table, see Fig. \ref{graphical-abstract}, was created by three domain experts  with several years of maintenance experience. While all previous steps in the workflow are fully automatic - and salable to other components - the look-up table will remain component specific and will always require a manual step.

\section{Evaluation} \label{Evaluation}


\subsection{Data Set Description} \label{Data Set Description}

The acquired data consists of $34$ irregularly sampled sensor signals in an industrial cooling system including numerous components with various data sources and a widely varying data domain, measured over a period of $18$ months. The cooling system is divided into numerous components, and each of them then again divided into several sub-components. Each sensor signal is assigned to a sub-component in the cooling system. We added $3$ additional timestamp signals including month, time (hour) and weekday to our observations resulting in a total of $37$ signals. By resampling the data to a regular frequency of $60$ seconds, we obtained a time series dataset of $37$-dimensional data samples. 
In general, the acquired sensor signals represent numerical values, e.g.\ measuring temperature, vibration or engine speed. However, the cooling system also includes signals of boolean values giving information about the system's health and up-counting signals giving information about the pause time of components in the system. 

In the course of digitization, sensors are often retrofitted or upgraded successively and additional sensors are integrated in the system posing challenges in the preprocessing and updating of the machine learning model. In general, the number of measured data points greatly increases over the measurement period of $18$ months for each sensor signal. For missing signals, we computed the mean and standard deviation and sampled from the corresponding normal distribution.


The dataset will be made public available, however, it will be anonymized due to privacy reasons. 

\subsection{Experimental Setup} \label{Experimental Setup}
Due to the steady increase in the number of data points in the system over $18$ months, data acquired in the first $10$ months is not representative and thus inadequate for testing. Therefore, we restricted our test data to the last $8$ months. We evaluated the machine learning workflow in two different cross validation approaches. In scenario $1$, we performed a conventional $4$-fold cross validation procedure on the last $8$ months (datasets $7$--$10$). In scenario $2$, we performed a $4$-fold cross validation approach on the last $8$ months (datasets $7$--$10$), including a basic training dataset consisting of the first $10$ months. We partitioned the data of the last $8$ months into $4$ equally sized folds, each fold receiving $54$ days of acquired data. For each fold $k$, we performed the following steps: Fold $k$ is held out for testing, and the remaining $3$ folds are used for training, in scenario $2$ the training data includes the basic dataset. The training data is preprocessed and prepared for modeling using the preprocessing steps described in section \ref{Preprocessing}. We then train the LSTM autoencoder described in section \ref{Anomaly Detection Model} for $50$ epochs with early-stopping on the training data. Finally, we test the autoencoder on the held-out test data. 

\begin{table}[th]
\small
	\caption{Organization of training and testing dataset in scenario $1$ and $2$ where $0$ refers to the training dataset and $1$ refers to the testing dataset}
	\begin{center}
		\begin{tabular}{ c|c|c|c|c|c|c|c|c} 
			\toprule[1.5pt]
		    & \multicolumn{4}{c|}{scenario 1} & \multicolumn{4}{c}{scenario 2} \\
		    \hline
		    round & 1 & 2 & 3 & 4 & 1 & 2 & 3 & 4 \\
		    \midrule[1.5pt]
		    dataset $1$ &  \multicolumn{4}{c|}{~} & $0$ & $0$ & $0$ & $0$ \\
		    \hline
		    dataset $2$ &  \multicolumn{4}{c|}{~} & $0$ & $0$ & $0$ & $0$ \\
		    \hline
		    dataset $3$ &  \multicolumn{4}{c|}{~} & $0$ & $0$ & $0$ & $0$ \\
		    \hline
		    dataset $4$ &  \multicolumn{4}{c|}{~} & $0$ & $0$ & $0$ & $0$ \\
		    \hline
		    dataset $5$ &  \multicolumn{4}{c|}{~} & $0$ & $0$ & $0$ & $0$ \\
		    \hline
		    dataset $6$ &  \multicolumn{4}{c|}{~} & $0$ & $0$ & $0$ & $0$ \\
		    \hline
		    dataset $7$ & $0$ & $0$ & $0$ & $1$ & $0$ & $0$ & $0$ & $1$ \\
		    \hline
		    dataset $8$ & $0$ & $0$ & $1$ & $0$ & $0$ & $0$ & $1$ & $0$ \\
		    \hline
		    dataset $9$ & $0$ & $1$ & $0$ & $0$ & $0$ & $1$ & $0$ & $0$ \\
		    \hline
		    dataset $10$ & $1$ & $0$ & $0$ & $0$ & $1$ & $0$ & $0$ & $0$ \\
		    \bottomrule[1.5pt]
	    \end{tabular}
		\label{tab1}
	\end{center}
\end{table}

\subsection{Ground Truth provided by Automatically Created Labels} \label{Ground Truth}
\begin{figure}[ht!]
	\centerline{\includegraphics[width=0.5\textwidth]{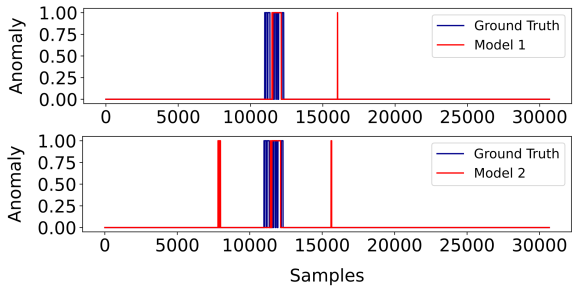}}
	\caption{Comparison of anomalies (healthy -- $0$, anomalous -- $1$) defined by ground truth or found by model over a time period of $8$ months, with models trained in scenario $1$ and $2$}
	\label{Anomalies}
\end{figure}

\begin{figure}[ht!]
	\centerline{\includegraphics[width=0.5\textwidth]{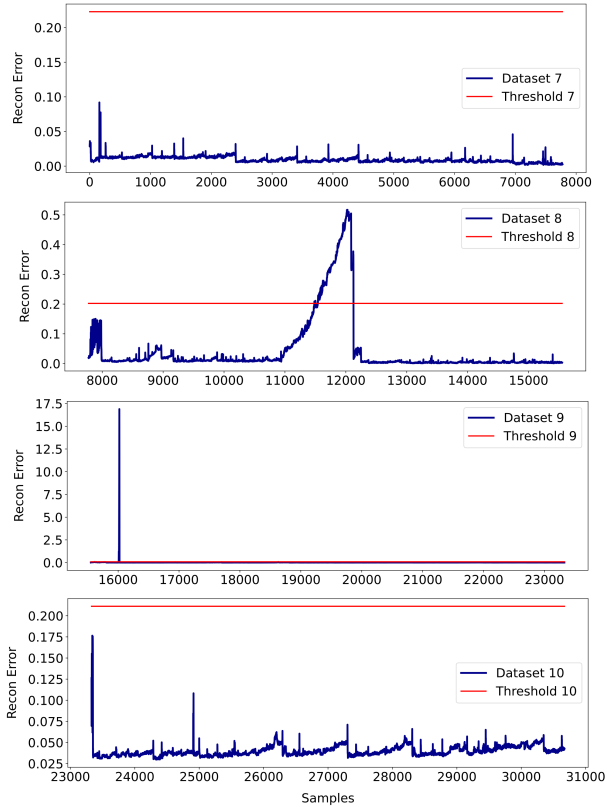}}
	\caption{Anomaly Detection in Scenario 1: total reconstruction error on test datasets $7$--$10$, each dataset of $1.8$ months ($54$ days), with respective threshold}
	\label{Scenario_4_fold_cv}
\end{figure}

\begin{figure}[ht!]
	\centerline{\includegraphics[width=0.5\textwidth]{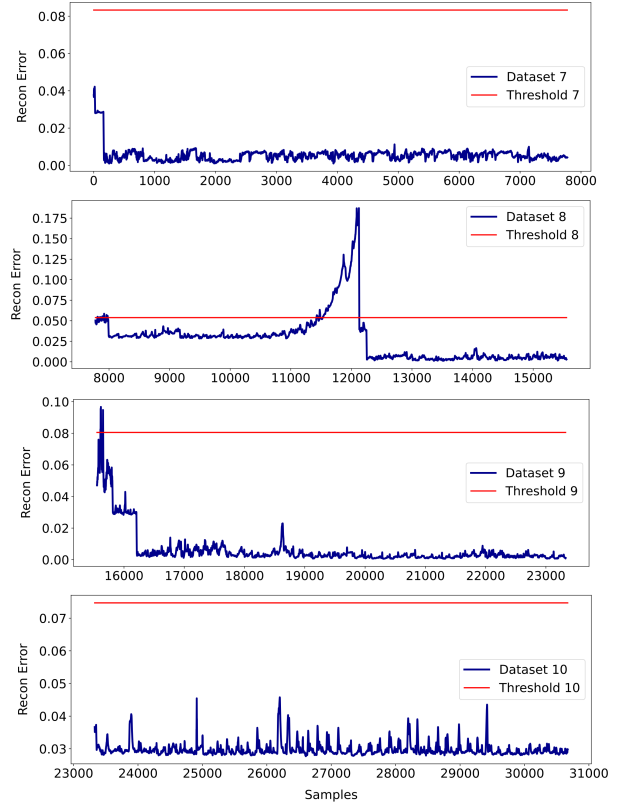}}
	\caption{Anomaly Detection in Scenario 2: total reconstruction error on test datasets $7$--$10$, each dataset of $1.8$ months ($54$ days), with respective threshold}
	\label{Scenario_10_fold_cv}
\end{figure}

We tested our proposed workflow for anomaly detection, fault localization and root cause analysis with thresholds based on expert knowledge, derived from PLC-system, failure log files, system sheets and documentation. A programmable logic controller (PLC) is a device used to control a machine or industrial system. The thresholds have been repeatedly evaluated in extensive feedback discussions with several domain experts and statistically confirmed. In case of absence of expert knowledge, we derived thresholds based on the $98$\% confidence interval. We obtained for each signal a threshold allowing us to provide automatically created labels. In the preprocessing procedure, after data cleaning, signal selection, resampling, missing values treatment but before data normalization, we compared the input features with the derived thresholds and obtained for each timestamp and signal a label (healthy -- $0$, anomalous -- $1$) enabling us to define a label for each sample 
\begin{align}
    F^{t} = \begin{pmatrix}
    \hat{x}^{1}_{t} &\hat{x}^{2}_{t} & \hdots & \hat{x}^{37}_{t} \\
    \hat{x}^{1}_{t+1} & \hat{x}^{2}_{t+1} & \hdots & \hat{x}^{37}_{t+1}\\
    \vdots & \vdots &   & \vdots\\
    \hat{x}^{1}_{t+9} & \hat{x}^{2}_{t+9} & \hdots & \hat{x}^{37}_{t+9} \end{pmatrix}  \in \mathbb{R}^{10 \times 37}.
\end{align}
We called  a timestamp $t+j$, $0 \le j \le 9$, anomalous if at least $10$ signals of all $37$ signals were anomalous at that timestamp, that is, at least $10$ of all values $\hat{x}^{1}_{t+j}, \hat{x}^{2}_{t+j}, \hdots ,\hat{x}^{37}_{t+j}$ exceed their thresholds. Finally, we called a sample $F_{t}$ with window size $w:=10$ anomalous if $t+j$ was an anomalous timestamp for all $0 \le j \le 9$. Then, we applied a smoothing filter to the labels, and labelled a sample $F_{t}$ anomalous if the smoothed value was greater or equal to $0.5$. Fig. \ref{Anomalies} shows the labels of data over a time period of $8$ months (datasets $7$--$10$ in Fig. \ref{Scenario_4_fold_cv} and Fig. \ref{Scenario_10_fold_cv}) and compares the results of the corresponding models to the automatically created labels (ground truth). 

\subsection{Evaluation Metrics} \label{Evaluation Metrics}
\begin{figure}[ht!]
	\centerline{\includegraphics[width=0.5\textwidth]{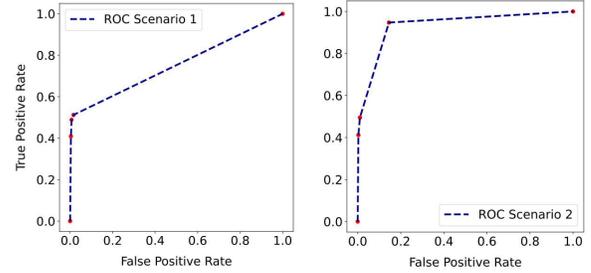}}
	\caption{ROC Curves based on 3 different thresholds for each scenario respectively}
	\label{ROC}
\end{figure}

In order to validate our model performance, we computed evaluation metrics including the F1 score, precision and recall for both scenarios, see Table \ref{tab2}. 
Further, we included ROC curves plotting the false-positive rate against the true-positive rate in several threshold settings, see Fig. \ref{ROC}. The true-positive rate is also known as sensitivity or recall. The ROC curves also helped us find the thresholds Eq. (\ref{threshold}). Inspired by the imaging domain, we used the Jaccard index to measure the similarity of machine learning and threshold derived anomalies. The Jaccard index measures the similarity of two datasets comparing their intersection and union.



\begin{table}[th]
\small
	\caption{Evaluation Metrics of Scenario 1 and Scenario 2}
	\begin{center}
		\begin{tabular}{ c|c|c|c } 
			\toprule[1.5pt]
		    & Scenario 1 & Scenario 2 & Ground Truth \\ 
		    \midrule[1.5pt]
	    	f1 score & $0.562$  & $0.527$ & \\ 
		    \hline
		    precision & $0.661$ & $0.564$ & \\ 
		    \hline
		    recall & $0.489$ & $0.495$ & \\ 
		    \hline
		    jaccard index & $0.391$ & $0.358$ & \\
		    \hline
		    consistency score & $0.920$ & $0.773$ & $0.619$ \\
		    \bottomrule[1.5pt]
	    \end{tabular}
		\label{tab2}
	\end{center}
\end{table}

\begin{figure}[ht!]
	\centerline{\includegraphics[width=0.5\textwidth]{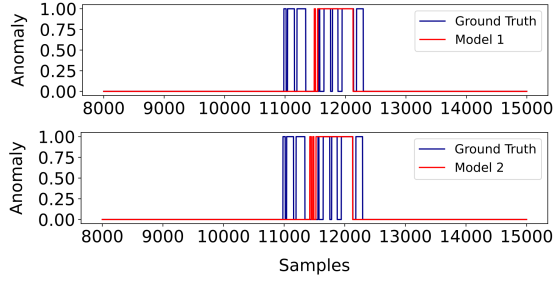}}
	\caption{Comparison of an anomaly (healthy -- $0$, anomalous -- $1$) defined by ground truth or found by model in dataset $8$ ($54$ days), with a model trained in scenario $1$ and $2$}
	\label{Part_Anomalies}
\end{figure}

Fig. \ref{Part_Anomalies} shows a part of Fig. \ref{Anomalies}. We use automatically created labels as ground truth references, however, these labels are also affected by precision and recall errors. We observe that the labels are not consistent in the occurrence of an anomaly. In order to measure the reliability of the system, we define a consistency score indicating the consistency of an anomaly over time, see Table \ref{tab2}. It is based on the assumption that, in the real-world, errors often persist consistently over a longer period of time.

Assuming an anomaly $a$ over a time period of several closed intervals $[s_{1}^{a}, e_{1}^{a}]$, $[s_{2}^{a}, e_{2}^{a}]$, $\dots$, $[s_{N}^{a}, e_{N}^{a}]$, motivated by Fig. \ref{Part_Anomalies}, we define the consistency score $\kappa_{a}$ for an anomaly $a$ as
\begin{align}
    \kappa_{a} := \frac{1}{e_{N}^{a} - s_{1}^{a}} \sum_{j=1}^{N} (e_{j}^{a} - s_{j}^{a}).
\end{align}
Further, we define the consistency score $\kappa$ for a model as 
\begin{align}
\kappa := \frac{1}{A} \sum_{a} (e_{N}^{a} - s_{1}^{a}) \kappa_{a}
\end{align}
where $A:= \sum_{a} (e_{N}^{a} - s_{1}^{a})$.


\subsection{Experimental Results}
We tested our workflow for anomaly detection, fault localization and root cause analysis described in section \ref{Fault Localization} and \ref{Root Cause Analysis} in scenario $1$ and scenario $2$. Scenario $1$ refers to the conventional $4$-fold cross validation procedure on the last $8$ months (datasets $7$--$10$). Scenario $2$ refers to the $4$-fold cross validation approach on the last $8$ months (datasets $7$--$10$) including a basic training dataset consisting of the first $10$ months. Fig. \ref{Scenario_4_fold_cv} shows the results for anomaly detection in scenario $1$. Fig. \ref{Scenario_10_fold_cv} shows the results for anomaly detection in scenario $2$. The plot shows the total reconstruction error Eq. (\ref{totalrecon}) and threshold computed according to Eq. (\ref{threshold}) on the respective test dataset over a period of $54$ days. Setting the sampling rate $r:=60$ (seconds) and the window size $w:=10$, we obtain $7776$ ($6 \times 24 \times 54$) data samples over a period of $54$ days. In both scenarios, the autoencoder detected an anomaly in dataset $8$ around data sample $12 000$ and an anomaly in dataset $9$ around data sample $16 000$. However, in scenario $2$, the autoencoder, trained on the training data including the basic dataset, detected the anomaly in dataset $9$ slightly earlier than in scenario $1$. Therefore, we suggest that including the data acquired in the first $10$ months increases the performance of the autoencoder.

\begin{figure}[ht!]
	\centerline{\includegraphics[width=0.4\textwidth]{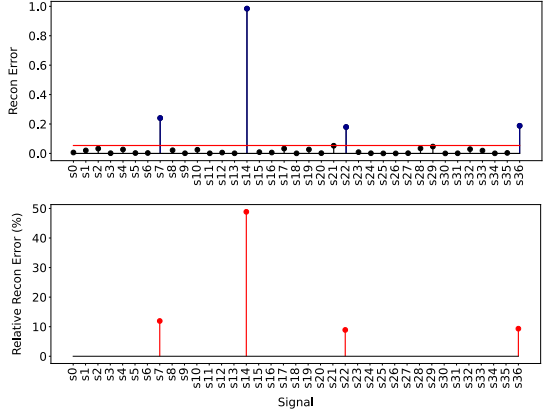}}
	\caption{Explained Fault Localization in Scenario 2: individual reconstruction error of anomalous data sample in dataset $8$}
	\label{RE_8}
\end{figure}

\begin{figure}[ht!]
	\centerline{\includegraphics[width=0.4\textwidth]{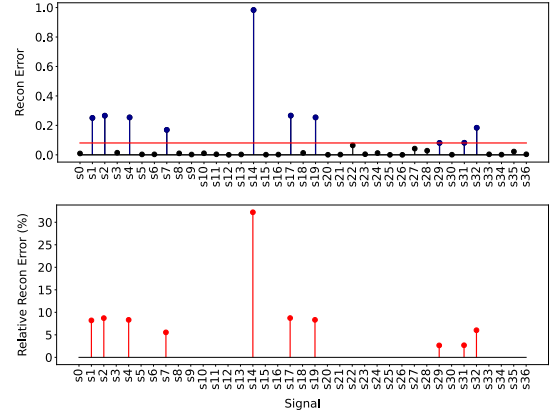}}
	\caption{Explained Fault Localization in Scenario 2: individual reconstruction error of anomalous data sample in dataset $9$}
	\label{RE_9}
\end{figure}

Fig. \ref{RE_8} shows the results for fault localization in scenario $2$ on dataset $8$. Fig. \ref{RE_9} shows the results for fault localization in scenario $2$ on dataset $9$. The plot shows the individual reconstruction error Eq. (\ref{indrecon}) of an anomalous data sample for all $37$ signals, absolutely and relatively with the respective threshold Eq. \ref{threshold}. The color indicates if the individual reconstruction error of signal $S_{i}$ exceeds the threshold.

\begin{table}[th]
\footnotesize
	\caption{Root Cause Analysis Scenario 2: component and failure type of anomalous signals of anomalous data sample in dataset $8$}
	\begin{center}
		\begin{tabular}{ c|c|c|c } 
			\toprule[1.5pt]
		    RE Contribution (\%) & Sensor & Component & Failure Type \\ 
		    \midrule[1.5pt]
	    	$11.94$ & Sensor $7$ & Component $3$ & $2$ \\ 
		    \hline
		    \textbf{48.86} & \textbf{Sensor 14} & \textbf{Component 1} & \textbf{1} \\ 
		    \hline
		    $8.91$ & Sensor $22$ & Component $1$ & $1$ \\ 
		    \hline
		    $9.33$ & Sensor $36$ & Component $13$ & $5$ \\  
		    \hline		     \multicolumn{4}{c}{Please note that the data is anonymized due to privacy reasons.} \\
		    \bottomrule[1.5pt]
	    \end{tabular}
	    \label{tab3}
    \end{center}
\end{table}

Table \ref{tab3} shows the results for root cause analysis in scenario $2$ on dataset $8$. Table \ref{tab4} shows the results for root cause analysis in scenario $2$ on dataset $9$. The explained fault localization and root cause analysis highlighted the most affected component in a consistent manner.

\begin{table}[th]
\footnotesize
	\caption{Root Cause Analysis Scenario 2: component and failure type of anomalous signals of anomalous data sample in dataset $9$}
	\begin{center}
		\begin{tabular}{c|c|c|c} 
			\toprule[1.5pt]
		    RE Contribution (\%) & Sensor & Component & Failure Type \\ 
		    \midrule[1.5pt]
	    	\textbf{8.21} & \textbf{Sensor 1} & \textbf{Component 1} &  \textbf{1} \\ 
		    \hline
		    \textbf{8.71} & \textbf{Sensor 2} & \textbf{Component 1} &  \textbf{1} \\ 
		    \hline
		    \textbf{8.33} & \textbf{Sensor 4} & \textbf{Component 1} & \textbf{1} \\ 
		    \hline
		    $5.55$ & Sensor $7$ & Component $3$ & $2$ \\
		    \hline
		    \textbf{32.22} & \textbf{Sensor 14} & \textbf{Component 1} & \textbf{1} \\ 
		    \hline
		    \textbf{8.73} & \textbf{Sensor 17} & \textbf{Component 1} & \textbf{1} \\ 
		    \hline
		    \textbf{8.33} & \textbf{Sensor 19} & \textbf{Component 1} & \textbf{1} \\ 
		    \hline
		    $2.66$ & Sensor $29$ & Component $10$ & $3$ \\ 
		    \hline
		    $2.69$ & Sensor $31$ & Component $12$ & $3$ \\ 
		    \hline
		    $6.03$ & Sensor $32$ & Component $12$ & $3$ \\ 
		    \hline
		    \multicolumn{4}{c}{Please note that the data is anonymized due to privacy reasons.} \\
		    \bottomrule[1.5pt]
	    \end{tabular}
		\label{tab4}
	\end{center}
\end{table}

\section{Conclusion and Future Work}

We provided a $18$ months dataset of multivariate time series data for an industrial cooling system including $34$ sensor signals and automatically created labels based on thresholds derived from expert knowledge and PLC-system. We presented our machine learning workflow for anomaly detection and explained fault localization suitable for multivariate and increasingly upgraded time series data. Therefore, we presented our preprocessing steps and provided an algorithm for explained fault localization and a root cause analysis enabled by integrated expert knowledge. 

We performed a conventional $4$-fold cross validation approach over a time period of $8$ months and a $4$-fold cross validation approach including a basic dataset of $10$ months and compared the model results to automatically created labels based on thresholds provided by domain experts. Using $4$-fold cross validation, we reached a F1-score of $0.56$, whereas the model results showed a higher consistency score (CS of $0.92$) compared to the automatically created labels (CS of $0.62$) -- indicating that the anomaly is recognized in a very stable manner. The automatically created labels, however, detected anomaly earlier. The main anomaly was found by the model and ground truth, and was also recorded in the log files. Further, the explained fault localization
highlighted the most affected component for the main anomaly in a very consistent manner.

A limitation of this work is the comparison of our model results to automatically created labels as ground truth references. These labels are also affected by precision and recall errors. Still, automatically created labels were the best available reference for this work and are frequently considered ground truth for other real-world applications. However, a strength of our study is that we also created and provided consistency scores indicating the consistency of an anomaly over time for autoencoder results and automatically created labels. We believe that this score will also be helpful in the future for evaluation using real-world data missing ground truth. It is based on the assumption that, in the real-world, errors often persist consistently over a longer period of time.

In the future, we would like to further expand root cause analysis and increase the transparency of our proposed workflow. Further, we aim to investigate whether the increase in reconstruction error and exceeding the anomaly thresholds can be predicted for the user. A further step is to bring the developed end-to-end model into a productive environment.  

Our work shows that scalable anomaly detection based on AI and explained fault localization is feasible for multivariate and increasingly upgraded time series data. 
Our proposed workflow provides a satisfying performance regarding the F1 score. We were also able to show that a root cause analysis enabled by integrated expert knowledge can be carried out without supervised learning highlighting the most affected component in a consistent manner. The integrated expert knowledge enabled ground truth references. We are sure that our results can also be transferred to other applications and will enable the monitoring of large systems in the future.


\section*{Data Availability Statement}
The dataset will be shared upon reasonable request. Please contact the senior author of the paper. Please cite this paper if you are using the dataset.

\section*{Acknowledgements}
We thank the Hauser experts their for the valuable contribution of domain knowledge, for their time and helpful discussions.

\section*{Funding}
This work was supported by the Austrian Research Promotion Agency (FFG) under Contract No. 883855.

\bibliographystyle{apacite}
\PHMbibliography{ijphm}

\end{document}